\def\year{2019}\relax
\documentclass[letterpaper]{article} 
\usepackage{aaai19}  
\usepackage{times}  
\usepackage{helvet}  
\usepackage{courier}  
\usepackage{url}  
\usepackage{graphicx}  
\usepackage{color}
\usepackage{subfig}
\usepackage[ruled,vlined]{algorithm2e}
\usepackage{amsmath}
\begin{document}
%
\title{Learning to Interpret Satellite Images Using Wikipedia}
\author{\\
\begin{tabular}{ccc}
\begin{tabular}{c} Evan Sheehan \\ esheehan@cs.stanford.edu \\ Stanford University \end{tabular}
&
\begin{tabular}{c} Burak Uzkent \\ buzkent@cs.stanford.edu \\ Stanford University \end{tabular}
&
\begin{tabular}{c} Chenlin Meng \\ chenlin@stanford.edu \\ Stanford University \end{tabular}
\end{tabular}
\\
\\
\begin{tabular}{cccc}
\begin{tabular}{c} Zhongyi Tang \\ zztang@stanford.edu \\ Stanford University \end{tabular}
&
\begin{tabular}{c} Marshall Burke \\ mburke@stanford.edu \\ Stanford University \end{tabular}
&
\begin{tabular}{c} David Lobell \\ dlobell@stanford.edu \\ Stanford University \end{tabular}
&
\begin{tabular}{c} Stefano Ermon \\ ermon@cs.stanford.edu \\ Stanford University \end{tabular}
\end{tabular}
}

\maketitle
\begin{abstract}
Despite recent progress in computer vision, fine-grained interpretation of satellite images remains challenging because of a lack of labeled training data.
To overcome this limitation, we propose using Wikipedia as a previously untapped source of rich, georeferenced textual information with global coverage. 
We construct a novel large-scale, multi-modal dataset by pairing geo-referenced Wikipedia articles with satellite imagery of their corresponding locations. To prove the efficacy of this dataset, we focus on the African continent and train a deep network to classify images based on labels extracted from articles.
We then fine-tune the model on a human-annotated dataset and demonstrate that this weak form of supervision
can drastically reduce the quantity of human-annotated labels and time required for downstream tasks. 
%
\end{abstract}

\section{Introduction}

Deep learning has been the driving force behind the recent improvements in fundamental computer vision tasks, including image classification, image segmentation, object detection and tracking, etc.~\cite{russakovsky2015imagenet,lin2014microsoft,han2018advanced}. These deep models, however, require training on high quality, large-scale datasets, and building these datasets is typically very costly. 
In particular, overhead, satellite images are specifically difficult and expensive to label because of humans' unfamiliarity with aerial perspectives~\cite{christie2018functional}.

One effective way to reduce the amount of training data needed is to perform pre-training on an existing, previously annotated dataset, such as ImageNet~\cite{deng2009imagenet}, and transfer the learned weights to the domain of interest \cite{raina2007self,dai2009eigentransfer}. 
However, the success of this approach diminishes if the underlying distributions and/or compositions of the pre-training and target datasets are not sufficiently similar. Such a problem is exceptionally pronounced in the satellite imagery space, as the entire frame of reference and perspective of an aerial image is altered compared to a natural image.
This has the unfortunate effect of rendering natural image datasets, such as ImageNet, less useful as pre-training mechanisms 
for downstream computer vision tasks in the satellite domain~\cite{pan2010survey}. 
\begin{figure*}[h]
\centering
\includegraphics[width=0.95\textwidth]{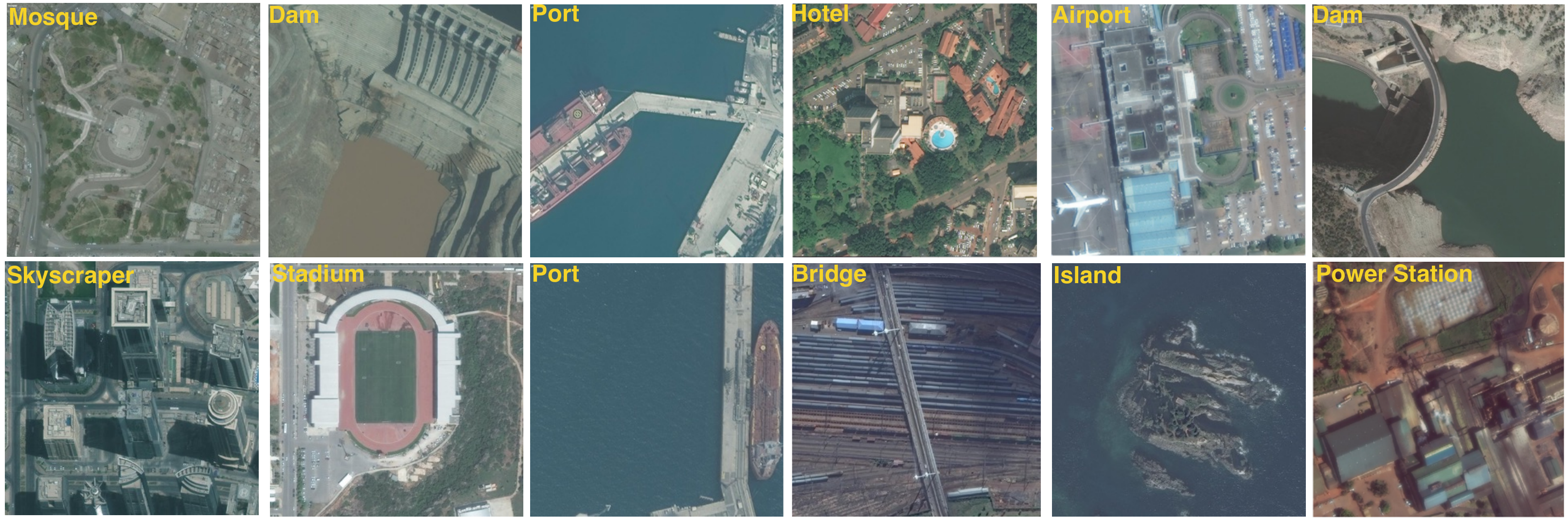}\\
\caption{Some of the satellite images labeled with Wikipedia articles. Zooming-in is recommended to visualize features.}
\label{fig:labeling_examples}
\end{figure*}



Because direct annotation is expensive, researchers have considered many creative ways to provide supervision without explicit labels. These include unsupervised~\cite{kingma2014semi}, 
label-free 
\cite{DBLP:journals/corr/abs-1805-10561,stewart2017label}, and weakly supervised learning methods~\cite{ratner2017snorkel}. A particularly effective strategy is to leverage co-occurrence statistics in a dataset, e.g.,  predict the next frame in a video, a missing word in a sentence \cite{DBLP:journals/corr/abs-1301-3781}, encourage a model to learn similar representations in the nearby satellite images \cite{jean2018tile2vec}, 
or predict relationships between entities such as images and text co-occurring together.
%
%
For example, leveraging co-occurring images and hashtags, \citeauthor{mahajan2018exploring} build a very large scale image recognition dataset consisting of more than 3 billion Instagram images across 17000 weakly-supervised labels obtained from textual hashtags and their WordNet \cite{miller1995wordnet} synsets. After pre-training on this extremely large dataset, they report $84\%$ top-1 accuracy on the downstream ImageNet classification task, equating to a $5\%$ improvement over the same model trained from scratch on ImageNet. 


Inspired by their work, we construct a new multimodal dataset by pairing  geo-referenced Wikipedia articles with satellite images corresponding to the article's location. 
To the best of our knowledge, this is the first time that Wikipedia has been used in conjunction with satellite images, and with 2.4 million potential article-image entries (if we were to collect all the images), 
our approach yields the \emph{largest labeled satellite image dataset ever assembled}. Additionally, 
as shown in Fig.~\ref{fig:article_distribution_world}, our approach is the first to provide comprehensive coverage of the \emph{entire world}.
By treating an article as an information rich label, we obtain highly detailed physical and qualitative context for each image.
For example, the article of Kampala\footnote{https://en.wikipedia.org/wiki/Kampala} contains excerpts such as ``\textit{Kampala was named the 13th fastest growing city on the planet, with an annual population growth rate of 4.03 percent by City Mayors. Kampala has been ranked the best city to live in East Africa,}''
highlighting the amount of information contained in Wikipedia articles.
Additionally, demographic, environmental, and social information is often readily available in structured form in many Wikipedia articles. Another exciting dimension that can be utilized to understand satellite images is the accessibility of Wikipedia articles over time, as using alterations in article content and distribution over time for a region could be helpful in detecting changes and predicting economic or population growth, etc.
We believe that the scale, coverage, and richness of this novel combination of crowdsourced annotations
and satellite images will enable new advances in computer vision and a wide range of new applications.
%

In this paper, 
we demonstrate the effectiveness of this approach for pre-training 
deep models for satellite image classification, as in~\cite{mahajan2018exploring}.
%
We label satellite images with curated summarization tags extracted from the article via an efficient, automated process.
For simplicity, we focus on a subset of 127000 images from the African continent, an area where ground labels are particularly scarce~\cite{jean2016combining}.
We then train a CNN architecture to predict these tags from their images.
This network is then evaluated on a downstream hand-labeled dataset, as in \cite{jean2018tile2vec}, where we prove that it attains more than $6\%$ higher accuracy compared to networks trained from scratch. We believe this novel combination of visual and textual information will enable new applications for research in the social sciences, economics, sustainability, etc., via machine learning, computer vision, and natural language processing. In particular, it will complement existing data sources from surveys and open data portals such as Open Street Maps (OSM), which typically lack global coverage and provide more coarse information. 

\section{Pairing Rich Crowdsourced Annonations from Wikipedia to Satellite Images}
\label{sect:dataset}
\subsection{Motivation}
\begin{figure}[h]
\centering
\includegraphics[width=0.50\textwidth]{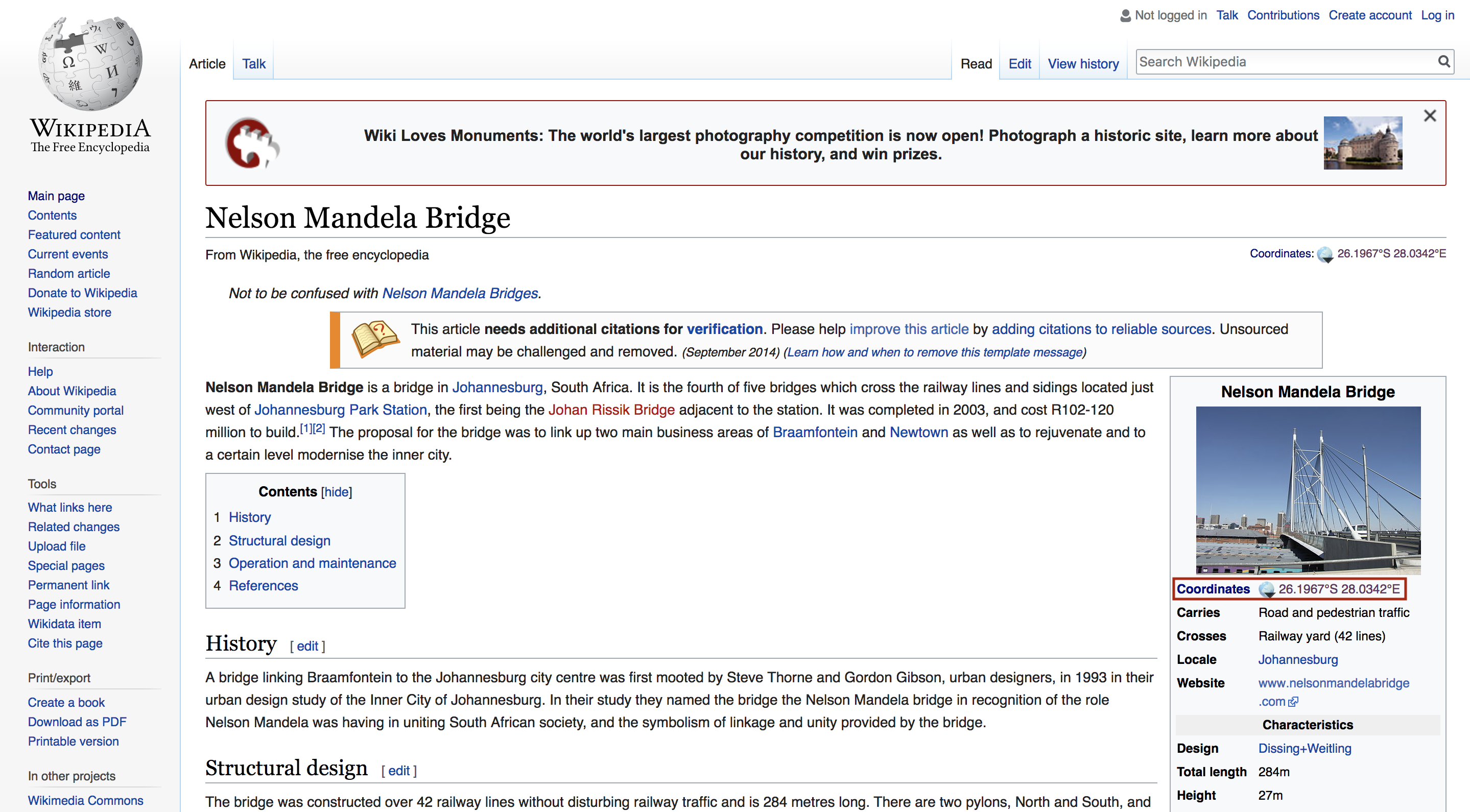}\\
\caption{An example Wikipedia article, with its coordinates boxed in red on the right-hand side. Observe the physical details about the bridge included in the article as well. Figure \ref{fig:labeling_examples} contains this article's paired ``Bridge'' image.}
\label{fig:article_example}
\end{figure}
Wikipedia is a large-scale, crowdsourced database spanning 302 languages
with over 47 million articles for numerous topics, entities, figures, events, etc~\cite{wiki:gen}. Of these 47 million articles, about 11$\%$ are contained by the English version. 
Out of these approximately 5 million
 articles, \emph{we found that roughly 1 million, or nearly 20$\%$, are geolocated}, meaning there is a latitude and a longitude associated with the article and implying it is possible to acquire a satellite image of its location from space (see Fig. \ref{fig:labeling_examples}).


Our key observation is that there is often a close relationship between the information in the  article and the visual content of the corresponding image.
Indeed, we can think of the article as an extremely detailed ``caption'' for the satellite image, providing an often-comprehensive textual representation of the satellite image, or an \emph{information rich label}. 
%
%
This label often contains structured data in the form of tables, called infoboxes, as well as raw text, allowing for the extraction of information about the physical state and features of the entity (e.g., elevation, age, climate, population). For example, text can be extracted from Fig. \ref{fig:article_example}'s infobox indicating that the ``Nelson Mandela Bridge'' is 284\textit{m} long, 27\textit{m} wide, and crosses over 42 railroad lines.
However, information stored in the less-structured body section of the article may be much more difficult to process and may require Natural Language Processing tools to extract at scale. Also, not all geolocated articles, such as those about historical battles or events, have a meaningful relationship with their corresponding satellite images.

\subsection{Collecting Wikipedia Articles}
The initial step for accomplishing our defined goal involved the acquisition of an English Wikipedia data dump of articles\footnote{https://dumps.wikimedia.org/enwiki/}, though, in the future, we plan to explore non-English Wikipedias to supplement the dataset. 
\begin{figure}[!t]
\centering
\includegraphics[width=0.50\textwidth]{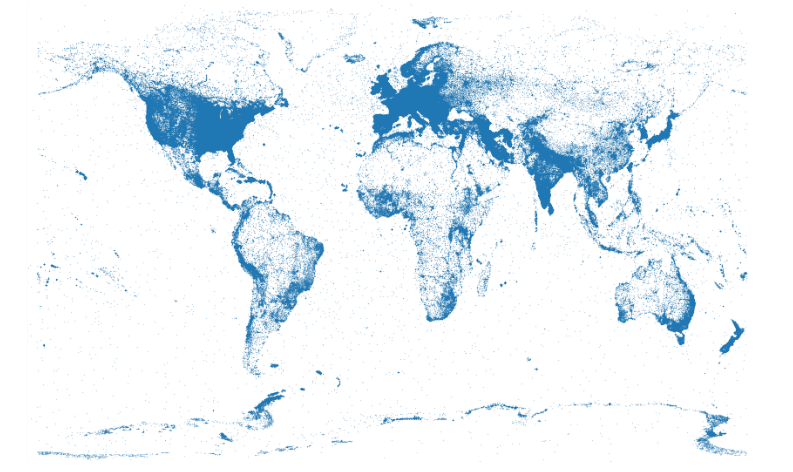}\\
\caption{Scatter plot of all geolocated Wikipedia articles. The article coordinates are drawn on a figure \textit{without any basemap}. The familiar continents' shapes emerge because of the inherent spatial distribution of the articles.
}
\label{fig:article_distribution_world}
\end{figure}
A Wikipedia article dump is stored as one large XML file containing all standard 
articles as well as numerous technical article stubs (e.g., internal communication pages, page redirects, etc.).
Thus, in order to analyze each relevant article individually, we first parsed the XML file into its constituent articles, netting roughly 5 million standard, titled articles. 
To isolate 
those which were geolocated, we then iterated through these 5 million articles and used regular expressions to find strings matching one of the archetypal coordinate patterns, such as:
\[\text{(1) }coord|lat|lon|display=title\]
\[\text{(2) for $d$} \in \{N,S,E,W\}:\]
\[coord|deg|min|sec|d|deg|min|sec|d|display=title\]
This resulted in the acquisition of 1.02 million articles possessing coordinates. 

The global distribution of these articles can be seen in Fig. \ref{fig:article_distribution_world}, plotted by their \textit{lat-lon}. Of interest is the scatter plot's precise definition of geographic features, especially coasts, \textit{without being overlaid on any map}, and its reflection of underlying socioeconomic patterns. For example, South Africa and the eastern USA are, respectively, more complete than the Sahara Desert and parts of the central/western USA, facts which reflect article density's likely correlation with geographic remoteness and population density. Also of interest is the apparent 
abundance of articles in certain regions, such as Syria, Iran, and North Korea, where 
OSM contains sparse data~\cite{mnih2012learning,maggiori2017convolutional,audebert2017joint}.
This global article coverage will likely improve even more with the inclusion of non-English Wikipedias as well.


\subsection{Pairing Articles with Satellite Images}
There are numerous ways to pair this global geolocated article dataset with satellite images. First, if the intended application of the dataset has a temporal dimension, the dates on which the satellite images were taken could be synced with temporal information from the articles, when available. This would maximize the temporal facet of Wikipedia by ensuring the image - article pairings were as correlated as possible. 
Applications with low noise thresholds might examine obtaining multiple images across multiple dates and satellites to reduce cloud cover issues.
Additionally, different applications will likely require images of different sizes, capturing different amounts of spatial context around the article's coordinates and reflecting confidence levels into the accuracy of the coordinates.
Environmental applications, e.g., studying heat emission from power plants, might require
infrared or multispectral satellite images of their locations to detect temperature changes, in addition to optical bands~\cite{kim2006satellite}. Similarly, research into the diffusion of religious buildings 
might pair information from the articles with high-resolution, small ground sample distance (GSD) images.


\section{Weakly Supervised Satellite Image Classification}
\begin{figure*}[h]
\centering
\includegraphics[width=0.95\textwidth]{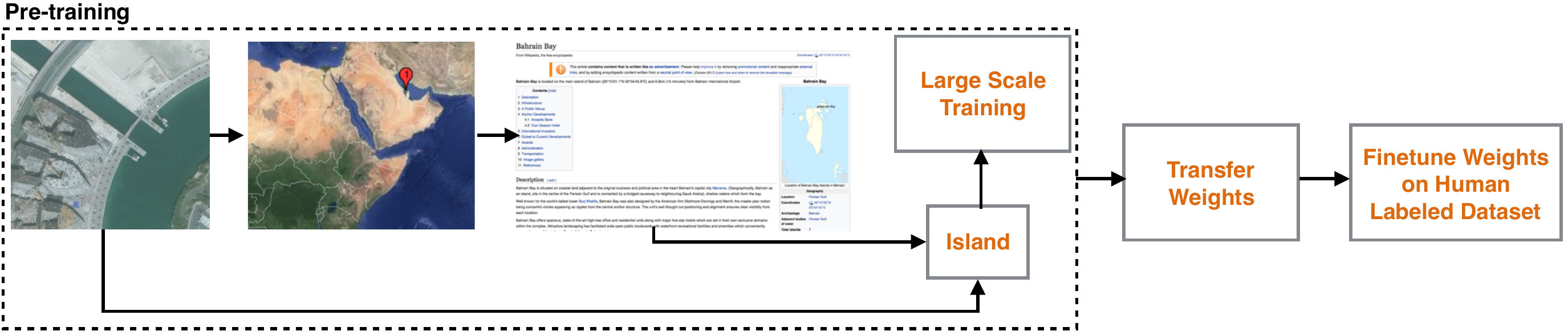}\\
\caption{The workflow of the proposed method: (1) Extract labels from articles using our labeling pipeline. (2) Match articles with images of their coordinates. (3) Pre-train on a large-scale dataset using the Wikipedia labels. (4) Transfer learned weights to a down-stream task.} 
\label{fig:workflow}
\end{figure*}

Exemplifying the diverse application possibilities highlighted in the last section, we instantiate a general Wikipedia article - satellite image framework for the weak supervision of image classification models~\cite{ratner2017snorkel}. That is, we explore whether it is possible to learn how to interpret satellite images using knowledge extracted from Wikipedia articles (see Figure \ref{fig:workflow}).
%
%
Additionally, to keep things practical, we extract images from only a subset of about 60000 articles from the African continent, out of the 1.02 million geolocated articles distributed globally.  
We believe our proof-of-concept is proven especially robust and generalizable since it succeeds even while utilizing data from Africa, given this is one of the the most data sparse regions on earth both in general and in English Wikipedia. Also, it positions us well to apply our research to a social task (e.g., predicting health or socioeconomic outcomes using machine learning) in the future using NLP tools and the textually rich information from the articles.


\subsection{Extracting Weak Labels from Wikipedia Articles}


In order to transition the current state of our dataset, containing 1.02 million geolocated Wikipedia articles, into a form where it could be used to reliably and precisely label satellite images, we first needed to engineer a standardized, easily-interpretable labeling system for articles. In general, each Wikipedia article containing an infobox, or a sequence of key-value data pairs at the beginning of an article, also possesses a template term which specifies the fields the infobox possesses. This term, which we call a ``raw label", functions as a highly granular and disparate label for an article (e.g. ``communes de france", ``new york train station") assigned by whoever authored the article. Using regular expressions, we were able to extract a raw label from $80\%$ of all geolocated articles since they appear in the same position, at the head of the infobox, for each article; the 20$\%$ we were unable to extract a raw label from did not possess infoboxes. When aggregated, we found there to be 978 unique raw labels 
across the 1.02 million geolocated articles. 

\begin{figure*}[t]
\centering
\includegraphics[width=0.95\textwidth]{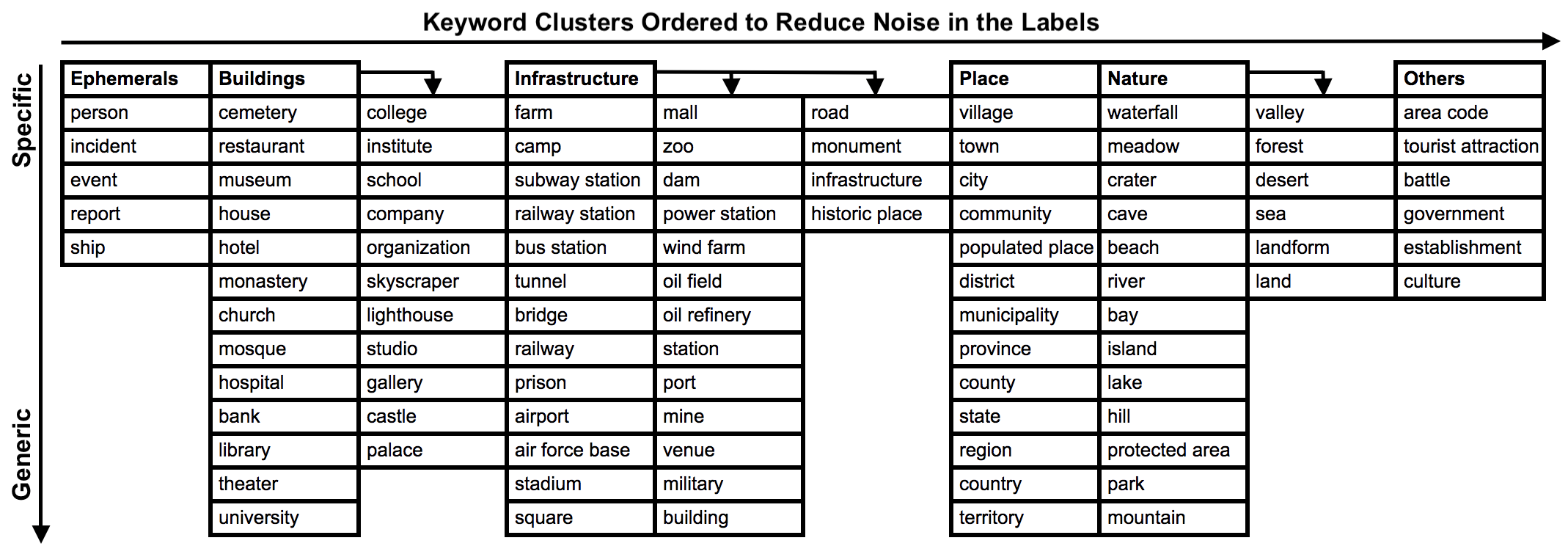}\\
\caption{The keyword hierarchy for labeling articles, searching each column top to bottom beginning with the leftmost column. 
}
\label{fig:curated label hierarchy}
\end{figure*}

Subsequently, to reduce this label space to a manageable size, we 
manually created a sequence of 97
keywords $L$ which we observed appeared often throughout the raw labels and also represented classes of entities discernible via satellite imagery. If the size of $L$ is not large enough or its keywords are not detailed enough, label noise increases dramatically and labels are able to be extracted for significantly fewer articles. Also, the creation of detailed labels allows for the application of the dataset to other domains (e.g. social sciences). $L$ is displayed in Fig. \ref{fig:curated label hierarchy}. 
After this, we clustered similar keywords in $L$ into 6
different broad categories, or clusters - ``Ephemeral'', ``Buildings'', ``Infrastructure'', ``Place'', ``Nature'', and ``Others'' -
and placed them in an order such that they 1) filtered out articles which fell into labels not detectable via satellite imagery (in the case of ``Ephemeral'' and ``Other'') and 2) minimized noise in the labels, since the terms in one cluster are searched for before the terms in the next cluster. Experimental creation and ordering of these clusters with such goals in mind led to the final sequence shown in Fig. \ref{fig:curated label hierarchy}, with ``Nature'' following ``Place'', etc. Within each cluster, we hierarchically structured the terms such that they progressed from specific entities (e.g. ``airport'') to generic ones (e.g. ``building''), with terms possessing the same level of specificity (e.g. ``bank'' and ``library'') randomly ordered among themselves.

To then label an article, we began with the leftmost cluster in the hierarchy, iterated through each keyword term within it, from top to bottom, and searched for the keyword in the article's raw label. If we found the term, we assigned it as the article's label. If not, we iterated through all the ``Category'' hyperlinks the article contained, which Wikipedia uses to group  similar articles together and most articles possess at the end of their bodies, and searched for the term in each one. If we discovered the term, then we set it as the article's label; if not, we repeated the entire process for the next term. This continued for all 97 possible keywords, moving to the next term in the cluster until the cluster was exhausted and then continuing on to the next cluster to the right. If, after trying all 97, no term was matched, we left the article unlabeled. Using this pipeline, 
we obtained labels for $96\%$, or 980000, of the 1.02 million geolocated articles across all 97 classes. Verifying this deterministic keyword labeling pipeline on a dataset of 500 hand-labeled articles revealed an accuracy of 94$\%$, with certain labels containing more noise than others.



\subsection{Label Verification with Neural Language Models}
During this labeling process, we also trained a Doc2Vec model \cite{le2014distributed} on the full Wikipedia article corpus and explored utilizing it to label articles and validate the label curation process previously described. 
Doc2Vec is an NLP method which takes in a corpus of documents $D$ and maps each document $d \in D$ to a vector $\mathbf{y}(d) \in R^n$ in an $n$-dimensional vector space, where $n$ is specified by the user. Documents which possess similar meanings are mapped to similar vectors, allowing a comparison to be made between any two documents in $D$. The model also maps each word it encounters in $D$ to an $n$-dimensional vector. In particular, it assigns a vector representation $\mathbf{y}(l)$  to each of the 97 keywords used in the previous section and shown in Fig. \ref{fig:curated label hierarchy}. 
In our experiments, we set $n$ to be 300.

In order to label a target article $d$ using this model, we attempted to directly compare $\mathbf{y}(d)$ with $\mathbf{y}(l)$ for all possible labels, picking the closest one. However, this approach was not successful, achieving less than $50\%$ top-1 accuracy. Instead, we propose an approach where we use the (weighted) average similarity between $\mathbf{y}(l)$ and the vector representation of all the words in  $\mathbf{y}(d)$. Specifically, we partition an article $d$ into $5$ sections: its body, category, hyperlinks, etc. For each section, we compute the average (cosine) similarity between $\mathbf{y}(l)$ and all the words in the section, based on their Doc2Vec representations. We then define the score of each label $l$ as the weighted similarity between $l$ and each section, where the $5$ weights for the $5$ sections are learnable parameters, which allows us to upweight the importance of certain sections.
After iterating over all 97 candidates, the label \textit{l} with the highest score is assigned to the article.

\begin{figure}[h]
\centering
\includegraphics[width=0.48\textwidth]{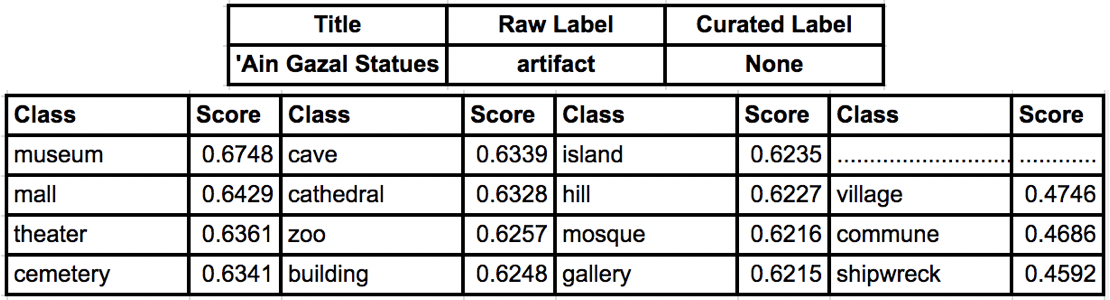}\\
\caption{Doc2Vec class scores for an article. Notice that our current pipeline is unable to extract a label for it (e.g. ``Curated Label''), but the model predicts numerous labels in its top 6 which are very close to its ground truth ``Raw Label'', ``artifact'', a class our keyword pipeline does not search for. 
}
\label{fig:doc2vec_example}
\end{figure}

For example, in Fig. \ref{fig:doc2vec_example}, the model labels the article ``'Ain Gazal Statues'' as ``museum'' since its label vector $\mathbf{y}(l)$ incurred the highest score in the labelling process. Even though our deterministic keyword pipeline fails to extract a label for this article, numerous labels in the top 6 it predicts (e.g. ``museum'', ``cemetery'', ``cathedral'') are quite similar to its ground truth (``artifact'').

The neural model performed well, achieving 95$\%$ top-5 accuracy and 70$\%$ top-1 accuracy in matching the label given via our deterministic pipeline. As demonstrated by Fig. \ref{fig:doc2vec_example},
the neural model was able to label articles which our deterministic pipeline had been unable to. We leave as future work the combination of the two approaches to improve accuracy and the direct use of distributed vector representations (such as the one in Fig. \ref{fig:doc2vec_example}) as weak labels.
\subsection{Acquiring Matching Satellite Imagery}
After assigning a curated label for $96\%$ of the 1.02 million geolocated articles (about 980000) using the previously described pipeline, 
we needed images to augment our dataset. While there are several ways to do this image-article pairing,
we utilized images of dimensions 1000$\times$1000 pixels and a ground sampling distance of 0.3\textit{m}. We also considered RGB and grayscale images in the dataset and didn't perform any filtering to remove cloudy images. The images in our Wikipedia dataset are the most recent images from DigitalGlobe satellites in our areas of interest. This is essential, as we utilized the most recent Wikipedia articles dump file.

\begin{figure}[h]
\centering
\includegraphics[width=0.30\textwidth]{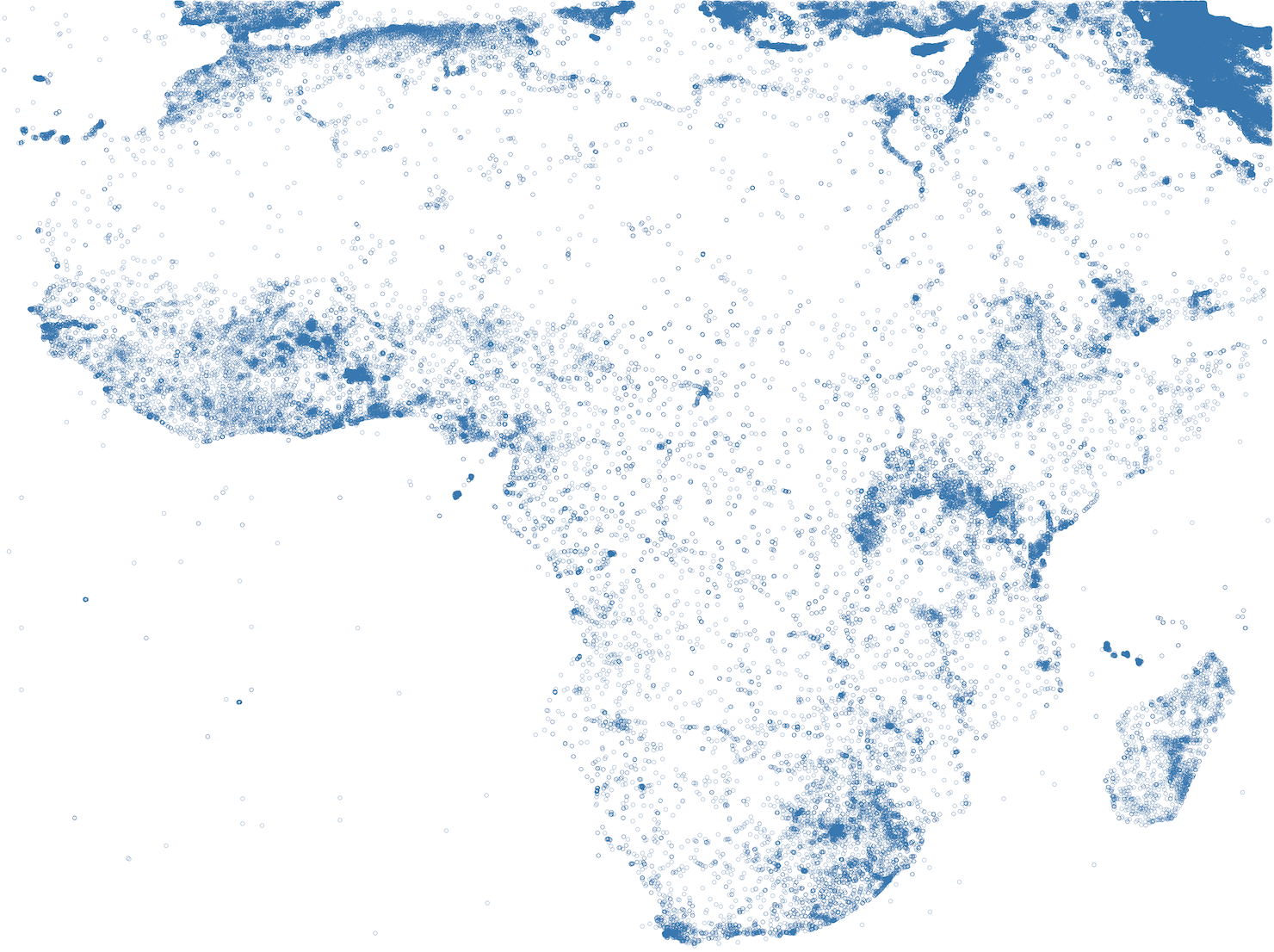}\\
\caption{Scatter plot of all images extracted from Africa, which a deep network is pre-trained on. \textit{No basemap is used.}}
\label{fig:article_distribution_africa}
\end{figure}

First, from the 980000 geolocated and labeled articles, we removed those which possessed 
labels that can not be represented by visual features, leaving us with roughly 958000 articles across 84 classes. Those classes removed included all labels in the ``Ephemeral'' and ``Other'' clusters as well as ``subway station'' and ``cave''. We then took a subset of 60000 African articles, which reduced the number of article classes to 71.
Then, for all classes we felt could be encompassed in a 300\textit{m}$\times$300\textit{m} area (e.g. \textit{building, school}, etc.), we acquired a satellite image centered at the coordinates of each article. For certain classes (e.g. \textit{road}, \textit{airport}, \textit{bridge}, \textit{oil field}, \textit{bay}, etc) which usually have key features beyond the 300\textit{m}$\times$300\textit{m} grid, we collected 9 images per article to make full use of the information. With the core image once again oriented on the article's coordinates, we overlapped multiple images by 150\textit{m} to cover a 600\textit{m}$\times$600\textit{m} area. For classes of natural objects, such as \textit{forest}, \textit{island}, \textit{sea}, etc., which contained an ``area'' field in their infobox, we obtained 300\textit{m}$\times$300\textit{m} images, overlapping by 150\textit{m} and totaling the minimum of either their ``area'' value or 2\textit{km}$^{2}$, surrounding the article's centroid coordinates. This process left us with 127000 images across 71 classes (Fig. \ref{fig:article_distribution_africa}). We would like to note, however, that applying this system globally would net 2.4 million images, distributed across 84 classes, according to our current pipeline.


We will publicly release the code we used to produce the dataset upon publication to encourage further research into jointly utilizing Wikipedia and  satellite images. 

\subsection{Post-processing Labels Considering Overhead Images}
\begin{figure}[h]
\centering
\includegraphics[width=0.40\textwidth]{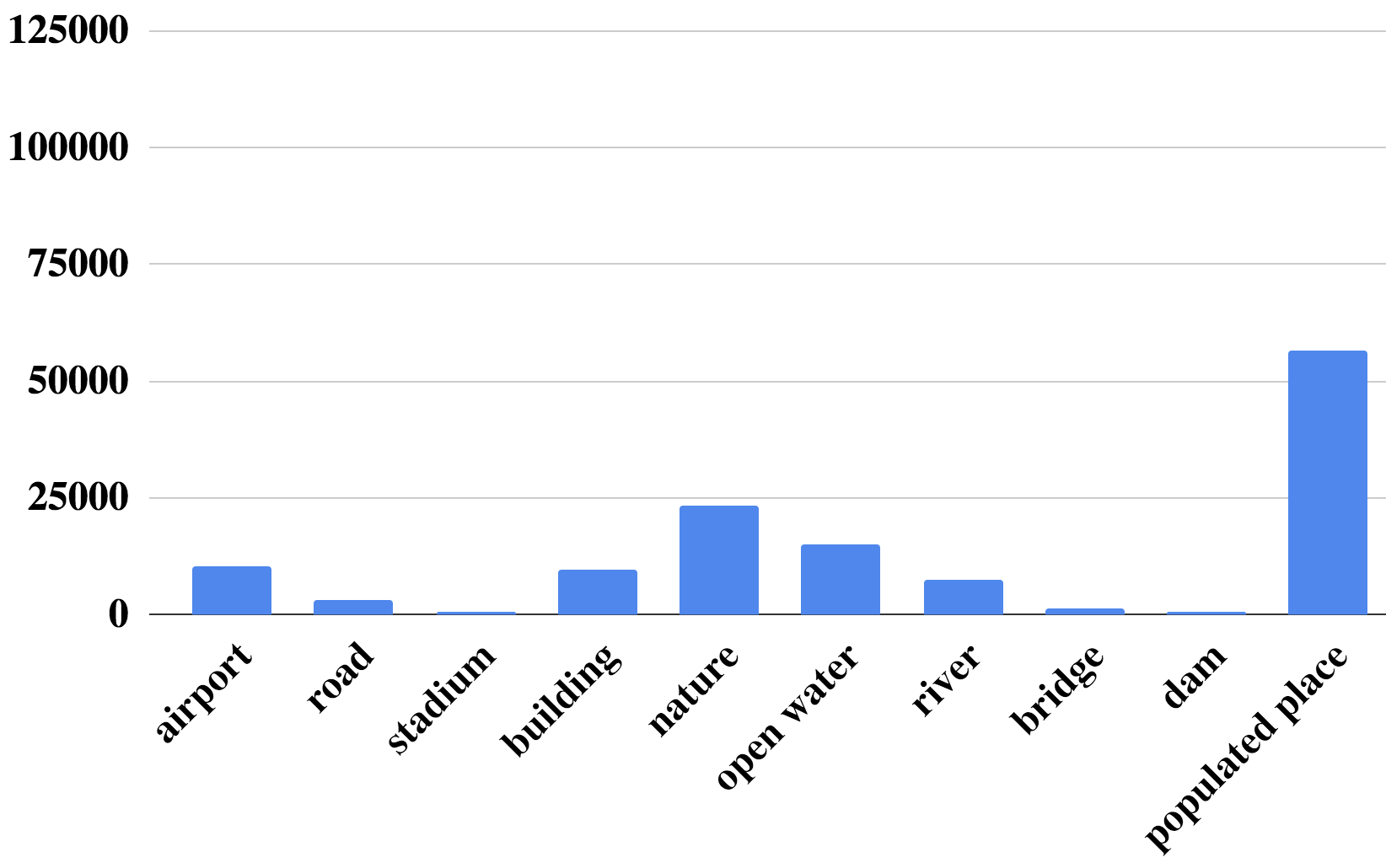}\\
\caption{African image distribution for the 10 merged-label version of the dataset.}
\label{fig:10 class histogram}
\end{figure}

After parsing and labeling the Wikipedia articles, we collected 127000 satellite images 
from Africa. Inspecting a sample of these images and their corresponding labels, we observed mutually-inclusive labels such as \textit{university}, \textit{college}, and \textit{school}. Although such objects are categorized into correct categories, it is almost impossible to tell the difference between them by looking at overhead images (e.g., \textit{university} vs. \textit{college}). Thus, to avoid having different labels represented by similar visual features, we engaged in manual merging operations. By iterating through a sample of images across all 71 labels, we determined which classes should go under each of the 10 general labels we created, shown in Fig.~\ref{fig:10 class histogram}. We experimented with different numbers of merged classes, starting from 2 and increasing to 14, but ultimately 10 labels outperformed the others slightly, while keeping the labels unchanged performed worse than all merged-label datasets. This verified that using mutually-inclusive labels would have created issues when training the network. One way to grapple with these mutually-inclusive labels is to design a hierarchical loss function that can tackle the mutually-inclusive labels as in \cite{redmon2016yolo9000}. Additionally, the similarities between the labels can be modeled using the Doc2Vec network, and the similarity scores can then be integrated into the loss function to avoid a large penalty when, for example, the model predicts \textit{school} for a \textit{university} target label, etc. We leave this as future work.

\section{Weak Supervision Experiments}
As the labels produced by our automatic procedure are inherently noisy, we evaluate the effectiveness of the dataset by pre-training deep models for satellite image classification. Specifically, we quantify the quality of the learned representations in a downstream classification task (via transfer learning).
To achieve this, we use a recently released large-scale satellite image recognition dataset named functional map of the world (fMoW)~\cite{christie2018functional}. The fMoW dataset consists of both multispectral and RGB images and contains about 83000  unique bounding boxes from large satellite images representing labeled objects. It also comes with temporal views from the same scenes. One can get more than 200000 samples by including temporal views in the dataset, however, they are not considered in this study as our goal is not to beat the benchmark but to prove the strength of the representations learned using information from Wikipedia. 

\subsection{Experimenting on the fMoW Dataset}

In our initial step, we pre-trained a deep convolutional network on our 127000 satellite image dataset, hoping to learn useful representations for down-stream tasks. The network used in our experiments was a 50-layer ResNet CNN \cite{he2016deep}, and we utilized the Adam optimization method in both our pre-training and fine-tuning experiments \cite{kingma2014semi}. During the pre-training stage, we tuned the learning rate and batch size to 0.001 and 64, respectively.

After training on our large-scale dataset, weakly labeled with Wikipedia articles, 
%
we fine-tune the network on a dataset with perfect annotations to verify that large-scale pre-training helps with the down-stream task. 
Because the fMoW dataset contains 83000 training images from 62 classes,
we experiment with different numbers of training images for the fine-tuning task to quantify the effectiveness of the representations learned in the pre-training stage. However, the validation dataset is left unchanged, and the pre-trained model fine-tuned on the fMoW dataset is compared to the same model trained from scratch on the fMoW dataset.
 

\subsubsection{Fine-tuning}
We use the ResNet50 model pre-trained on the Wikipedia dataset and transfer the weights for classification on the fMoW dataset. There are two classical approaches in fine-tuning a deep network: (1) training all layers, and (2) freezing all the layers other than the classification layer. Our experiments favor the first approach, and we only include results from the first method in the rest of the paper. The classification layer of the model, on the other hand, is changed to perform 62-way classification to match the number of labels in the fMoW dataset. Finally, resized, 224$\times$224 pixel RGB images are used as input to the model as in the pre-training task.

\subsubsection{Training from Scratch}
In this step, the ResNet50 model is trained from scratch on the fMoW dataset. All the hyper-parameters are kept similar to the model pre-trained on the Wikipedia dataset, following the practice suggested in \cite{mahajan2018exploring}. To be consistent with the pre-trained model, resized, 224$\times$224 pixel RGB images are used. 

\subsection{Results} In the first task, we perform experiments on the full fMoW training set and train a model from scratch and fine-tune the one pre-trained on the Wikipedia dataset. This way, the quality of the representations learned on the pre-training task can be quantified.  Table~\ref{table:large_scale_exp_on_fmow} shows the results on the test split of the fMoW dataset.
\begin{table}[h]
\centering
\resizebox{0.48\textwidth}{!}{\begin{tabular}{@{}l|c|c@{}}
Model & \begin{tabular}[c]{@{}c@{}}ResNet50 \\ (\textit{trained from scratch}) \end{tabular} & \begin{tabular}[c]{@{}c@{}} ResNet50 \\ (\textit{pre-trained} on Wikipedia) \end{tabular} \\
\hline
\begin{tabular}[c]{@{}c@{}}Accuracy \\ (\textit{66000 Samples}) \end{tabular} & 33.84 (\%) & 36.78 (\%) \\
\hline
\begin{tabular}[c]{@{}c@{}}Accuracy \\ (\textit{30000 Samples}) \end{tabular} & 27.23 (\%) & 33.38 (\%) \\
\hline
\begin{tabular}[c]{@{}c@{}}Accuracy \\ (\textit{20000 Samples}) \end{tabular} & 22.38 (\%) & 29.24  (\%)
\end{tabular}}
\caption{Top-1 accuracy performance on fMoW's test split.}
\label{table:large_scale_exp_on_fmow}
\end{table}


As seen in Table~\ref{table:large_scale_exp_on_fmow}, on a down-stream task with limited samples, the model pre-trained on the 127000 images from the Wikipedia dataset outperforms the one trained from scratch on the target task by more than $6\%$. As expected, the gap between two models increases as the number of training samples are decreased from 66000 to 20000. It should be highlighted that both models should improve with the use of further training data.

In our next experiment, we select classes that individually co-exist in both our Wikipedia and fMoW datasets so that the learned representations are tested purely on the classes that are represented at a reasonable scale in our pre-training task, as in \cite{mahajan2018exploring}. The goal of this task is to understand how much having similar classes in both pre-training and down-stream task helps in improving accuracy in the down-stream task.
\begin{table}[h]
\centering
\resizebox{0.48\textwidth}{!}{\begin{tabular}{@{}l|c|c@{}}
Model & \begin{tabular}[c]{@{}c@{}}ResNet50 \\ (\textit{trained from scratch}) \end{tabular} & \begin{tabular}[c]{@{}c@{}} ResNet50 \\ (\textit{pre-trained} on Wikipedia) \end{tabular} \\
\hline
\begin{tabular}[c]{@{}c@{}}Accuracy \\ (\textit{22000 Samples}) \end{tabular}  & 49.97 (\%) & 55.07 (\%) \\
\hline
\begin{tabular}[c]{@{}c@{}}Accuracy \\ (\textit{11000 Samples}) \end{tabular}  & 45.18 (\%) & 50.73 (\%)
\end{tabular}}
\caption{Top-1 accuracy performance on the test split of the fMoW dataset on 19 classes that are represented reasonably in our Wikipedia dataset. Both models are trained on either 22000 or 11000 images from 19 categories sampled from the train split of the fMoW dataset.}
\label{table:19_classes_experiments_fMoW}
\end{table}

As seen in Table~\ref{table:19_classes_experiments_fMoW}, pre-training on our Wikipedia dataset improves the accuracy by $6\%$ compared to the baseline model on our target task containing 22000 images from 19 categories. By comparing Table~\ref{table:19_classes_experiments_fMoW} to Table~\ref{table:large_scale_exp_on_fmow}, we can conclude that building a large-scale dataset that represents the classes existing in the target task is a better way of pre-training than the one where the target task contains classes unrepresented in the pre-training task at a reasonable scale.

Finally, we conduct experiments on three-way classification tasks by sampling three classes from the train split of fMoW. The class pairs selected in this case are also represented as individual classes in our Wikipedia dataset from Africa, with 683, 13543, and 14712 samples for \textit{stadium}, \textit{airport}, and \textit{lake} respectively.
\begin{table}[h]
\centering
\resizebox{0.48\textwidth}{!}{\begin{tabular}{@{}l|c|c@{}}
Model & \begin{tabular}[c]{@{}c@{}}ResNet50 \\ (\textit{trained from scratch}) \end{tabular} & \begin{tabular}[c]{@{}c@{}} ResNet50 \\ (\textit{pre-trained} on Wikipedia) \end{tabular} \\
\hline
\begin{tabular}[c]{@{}c@{}}Accuracy \\ (\textit{64 Samples}) \end{tabular} & 58.94 (\%) & 82.03 (\%) \\
\hline
\begin{tabular}[c]{@{}c@{}}Accuracy \\ (\textit{500 Samples}) \end{tabular}& 71.09 (\%) & 86.72 (\%) \\
 \hline
\begin{tabular}[c]{@{}c@{}}Accuracy \\ (\textit{1200 Samples}) \end{tabular} & 82.81 (\%) & 90.63 (\%)
\end{tabular}}
\caption{Top-1 accuracy performance on the test split of the fMoW dataset on three classes (\textit{airport}, \textit{stadium}, \textit{lake$\_$or$\_$pond}) that are represented reasonably in our Wikipedia dataset. Even with a small number of training samples, the model pre-trained on Wikipedia dataset can achieve reasonable performance.}
\label{table:three-way-classification}
\end{table}

As seen in Table~\ref{table:three-way-classification}, the ResNet50 model pre-trained on our Wikipedia-labeled image recognition task outperforms the same model trained from scratch by a large margin. As an extreme case, we only used 64 samples from three classes to fine-tune the pre-trained model. With about $82\%$ top-1 accuracy, the pre-trained model performs similarly to the baseline ResNet50 model trained on the 1200 samples. On the other hand, as expected, the performance of the baseline model improves with an increase in the number of training samples. These results clearly indicate that weakly supervised training using Wikipedia labels can be highly beneficial for substantially reducing the amount of human labeled data required in down-stream tasks.

\subsubsection{Classification on fMoW without Fine-tuning} Apart from fine-tuning and testing the model on the fMoW dataset, one can also test the learned representations in the pre-training stage without any fine-tuning in the down-stream task. In this direction, we choose a number of classes that co-exist in the Wikipedia and fMoW datasets. These classes are \textit{airport}, \textit{bridge}, \textit{dam}, \textit{stadium}, \textit{lake}, and \textit{port}. In the training stage, we exclude samples from the other classes in the African Wikipedia dataset and train the model on only samples from these six classes. The network is then tested on the fMoW samples from the same classes without doing any fine-tuning on the train split of the fMoW dataset. By using only weak supervision considering Wikipedia labels from Africa, we obtained $53.4\%$ accuracy on the fMoW dataset containing samples from all around the world. We expect that scaling up our approach to collect images worldwide would further boost the accuracy on the fMoW dataset.


\section{Conclusion}
In this study, we proposed a novel combination of satellite images and crowdsourced annotations from georeferenced Wikipedia articles.
To the best of our knowledge, this is the first time that Wikipedia has been used this way. Our approach yields a large scale, multimodal dataset combining rich visual and textual information for millions of locations all over the world, and including additional languages beyond English will likely improve coverage even more.
In this paper, we use rather crude information automatically extracted from the articles to teach a neural network how to interpret the corresponding satellite images, and show that this pre-training step leads to accuracy improvements on standard benchmark tasks.
More broadly, we believe this approach will open new research avenues both in computer science and the social sciences. In this vein, fine-grained spatial information extracted from the articles could complement existing data sources, especially in the developing world where data is typically very scarce. 

\bibliography{references}
\bibliographystyle{aaai}
\end{document}